\documentclass[runningheads]{llncs}
\usepackage{cite} 
\usepackage{graphicx}
\graphicspath{{./figures/}}
\usepackage{color}
\usepackage{amsmath}
\usepackage[normalem]{ulem}
 
\usepackage{multirow}
\usepackage{booktabs}
\usepackage{subfig}
\usepackage{caption}

\newcommand{\keywords}[1]{\par\addvspace\baselineskip
\noindent\keywordname\enspace\ignorespaces#1}

\begin{document}
\mainmatter  
\title{ASCNet: Adaptive-Scale Convolutional Neural Networks for Multi-Scale Feature Learning}
\titlerunning{  }
\author{Mo Zhang\inst{1\dagger}, Jie Zhao\inst{2\dagger}, Xiang Li\inst{3\dagger}, Li Zhang\inst{2\ddagger}, Quanzheng Li\inst{2,3,4\ddagger}}
\institute{$^{1}$Center for Data Science, Peking University, Beijing 100871, China; $^{2}$Center for Data Science in Health and Medicine, Peking University, Beijing 100871, China; $^{3}$Center for Clinical Data Science, Masschusetts General Hospital, Harvard Medical School, Boston MA 02115, USA;  $^{4}$Laboratory for Biomedical Image Analysis, Beijing Institute of Big Data Research, Beijing 100871, China; \\
   \  $\dagger$ Joint First Authors, $\ddagger$ Joint Corresponding Authors}
\authorrunning{  }

\maketitle
\begin{abstract}
Extracting multi-scale information is key to semantic segmentation. However, the classic convolutional neural networks (CNNs) encounter difficulties in achieving multi-scale information extraction: expanding convolutional kernel incurs the high computational cost and using maximum pooling sacrifices image information. The recently developed dilated convolution solves these problems, but with the limitation that the dilation rates are fixed and therefore the receptive field cannot fit for all objects with different sizes in the image. We propose an adaptive-scale convolutional neural network (ASCNet), which introduces a 3-layer convolution structure in the end-to-end training, to adaptively learn an appropriate dilation rate for each pixel in the image. Such pixel-level dilation rates produce optimal receptive fields so that the information of objects with different sizes can be extracted at the corresponding scale. We compare the segmentation results using the classic CNN, the dilated CNN and the proposed ASCNet on two types of medical images (The Herlev dataset and SCD RBC dataset). The experimental results show that ASCNet achieves the highest accuracy. Moreover, the automatically generated dilation rates are positively correlated to the sizes of the objects, confirming the effectiveness of the proposed method.
\end{abstract}
\section{Introduction}
Medical image segmentation is challenging despite the substantial advance in deep convolutional neural networks (CNNs). To segment objects of different sizes, CNNs should be able to extract multi-scale information. However, due to the limited kernel size, the classic convolutional layer can only extract image information within a certain range (called the receptive field). Traditionally, two types of methods are proposed to enlarge the receptive field. One is to expand the convolutional kernel with substantially increased computational cost and time \cite{peng2017large}. The other introduces the maximum pooling operation, which roughly doubles the receptive field but loses part of image information \cite{long2015fully}.

Recent works aim to find better solutions to this problem. Yu et. al. \cite{yu2015multi,yu2017dilated} first propose to use the dilated convolution to increase the sizes of receptive fields without losing information. Two types of studies have further improved the method. The first is to arbitrarily assign the dilation rate for each convolutional layer. For example, in order to solve the problem of small and dense target segmentation, Hamaguchi et al. \cite{hamaguchi2018effective} and Li et al. \cite{li2018csrnet} use different configurations of dilation rates in the deep segmentation models and achieve high accuracies. The second category is to use a pyramid structure of dilated convolutions to fuse the information from multiple scales. Chen et al. develop a module called Atrous Spatial Pyramid Pooling (ASPP), in which multi-scale information is extracted by parallel convolutions with different dilation rates \cite{chen2018deeplab}. Mehta et al. \cite{mehta2018espnet} and Wang et al. \cite{wang2018understanding} also report similar pyramid (hybrid) dilated convolutional modules to improve the semantic segmentation. 

Although the aforementioned methods show that the dilated convolution is effective, it has two major limitations: 1) training a dilated CNN requires extra efforts on tuning the dilation rates; 2) manually designed dilation rate only provides a fixed-size receptive field, which may not be suitable for different objects in the same image. So following the idea of dilated convolution, we propose a new convolutional network architecture, referred as Adaptive-Scale Convolutional Network (ASCNet). By introducing a small 3-layer convolutional structure (called additional network), ASCNet adaptively learns the dilation rate for each pixel on the input feature map. Such pixel-level dilation rates form a dilation rate field, which is applied to all convolutional layers. The proposed network is optimized in an end-to-end training process, where the convolutions are able to extract image information of different objects in the optimal receptive fields.

We compare the classic CNN, the dilated CNN and the proposed ASCNet on both Herlev \cite{jantzen2005pap} and RBC datasets \cite{zhang2018rbc}. The experimental results show that ASCNet improves the accuracy of image segmentation on both datasets with a slight increment of computational cost. Furthermore, the dilation rates learned by the proposed ASCNet are positively correlated to the sizes of the objects from both datasets. It confirms that the proposed method effectively finds the optimal dilation rates for objects with different sizes.
\section{Method}
\subsection{Classic and dilated convolutions}
The classic convolutional kernel samples the input feature map on an integer grid, for example, the grid $\mathcal{R}$ for a $3\times3$ convolutional kernel is $\mathcal{R}=\{(-1,-1),(-1,0),\cdots,(1,0),(1,1)\}$. Classic convolution can be expressed as:
\begin{equation}
\textbf{y}(\textbf{p}_0)=\sum_{\textbf{p}_n\in \mathcal{R}}\textbf{w}(\textbf{p}_n)\cdot \textbf{x}(\textbf{p}_0+\textbf{p}_n),
\label{eq:classic_conv}
\end{equation}
where $\textbf{x}(\textbf{p}_0+\textbf{p}_n)$ denotes the input value at pixel $(\textbf{p}_0+\textbf{p}_n)$, $\textbf{y}(\textbf{p}_0)$ denotes the output value at pixel $\textbf{p}_0$, and $\textbf{w}(\textbf{p}_n) [\textbf{p}_n\in \mathcal{R}]$ is the weight at the offset $\textbf{p}_n$ in the kernel. The classic convolution has small receptive field because of the limited grid size, so it usually works with successive downsampling process to enlarge the receptive field at a cost of spatial resolution.

Dilated convolution \cite{yu2015multi,shi2017single} provides an alternative way to enlarge the receptive field by inserting ``holes'' into the convolutional kernel with a dilation rate of $r$, which can be defined as:
\begin{equation}
\textbf{y}(\textbf{p}_0)=\sum_{\textbf{p}_n\in \mathcal{R}}\textbf{w}(\textbf{p}_n)\cdot \textbf{x}(\textbf{p}_0+{r}\cdot \textbf{p}_n),
\label{eq:dilated_conv}
\end{equation}
where $r$ is an integer, so dilated convolution is a discrete operation and it degenerates into classic convolution when $r=1$. Compared to successive downsampling process, dilated convolution is able to systematically integrate multi-scale contextual features without losing spatial information.
\subsection{Adaptive-scale convolution (ASC)}
Following the idea of the dilated convolution, we design a novel convolutional module called adaptive-scale convolution (ASC). ASC has three distinctions from the dilated convolution: 1) the dilation rate $r$ is learned from raw data instead of manually assigned; 2) different pixels on a certain feature map employ adaptive dilation rates rather than the same one; 3) the dilation rate $r$ is a float value rather than an integer. 
\begin{figure}[!b]
	\centering
	\includegraphics[width =0.65\textwidth]{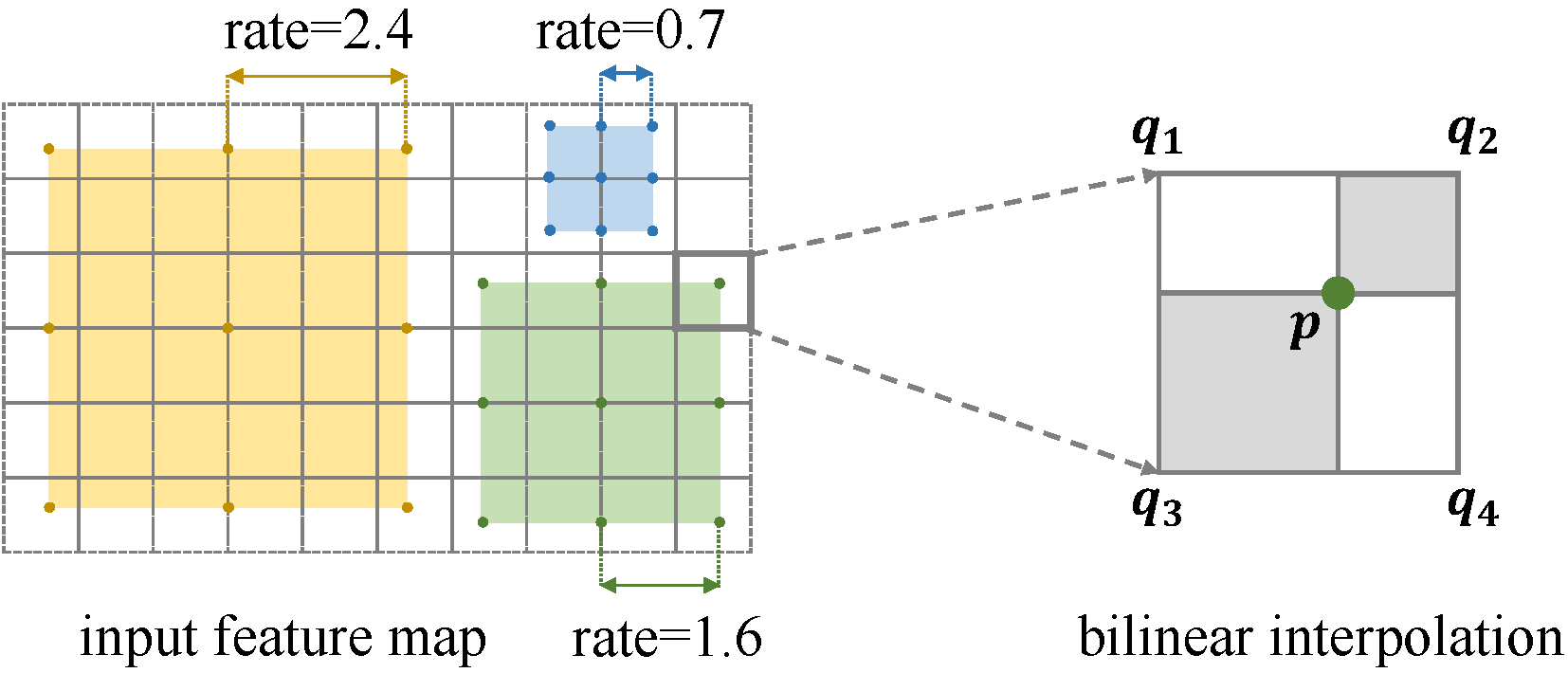}
	\caption{The sampling process of ASC module. In the left panel, the colored dots represent the sampling locations of the ASC kernel. In the right panel, a bilinear interpolation is adopted when the dilation rate is a float number.}
	\label{fig:ASC_effect}
\end{figure}

More specifically, we use an additional network to learn the pixel-level dilation rates, which form a rate field with one channel and the same size as the input feature map. Different dilation rates can thus be used on the feature map simultaneously, providing optimal receptive fields to objects with different sizes. The ASC can be expressed as:
\begin{equation}
\textbf{y}(\textbf{p}_0)=\sum_{\textbf{p}_n\in \mathcal{R}}\textbf{w}(\textbf{p}_n)\cdot \textbf{x}(\textbf{p}_0+{r(\textbf{x}_0,\theta)}\cdot \textbf{p}_n),
\label{eq:asc_conv}
\end{equation}
where $r$ is learned by the network with the input of raw image $\textbf{x}_0$ and the parameters of $\theta$.

As shown in Fig.~\ref{fig:ASC_effect}, the learned dilation rate in our proposed ASC is usually a float value and the sampling location $\textbf{p}~  (\textbf{p}=\textbf{p}_0+{r(\textbf{x}_0,\theta)}\cdot \textbf{p}_n)$ may not lie exactly on the regular grid. 
We therefore introduce a bilinear interpolation to compute the value $\textbf{x}(\textbf{p})$ at $\textbf{p}$ from the integer coordinates on the input feature map \cite{li2017dense},
\begin{equation}
\textbf{x}(\textbf{p})=\sum_{\textbf{q}}f_{int}(\textbf{q},\textbf{p})\cdot\textbf{x}(\textbf{q}),
\label{eq:bin_inter_len}
\end{equation}
where $\textbf{q}$ enumerates all integer locations on the input feature map, $f_{int}$ is the bilinear interpolation kernel defined as:
\begin{equation}
f_{int}(\textbf{q},\textbf{p})=max(0,1-|q_x-p_x|) \cdot max(0,1-|q_y-p_y|),
\label{eq:inter}
\end{equation}
where $p_x(p_y)$ and $q_x(q_y)$ denote $x$-($y$-)coordinate of $\textbf{p}$ and $\textbf{q}$, respectively. In fact, only the four integer locations $\textbf{q}_i [i=1,2,3,4]$ adjacent to $\textbf{p}$ will contribute to the interpolated value, because $f_{int}$ equals 0 on other locations. The right panel of Fig.~\ref{fig:ASC_effect} demonstrates an example of $\textbf{p}$ falling inside the grid. The Eq.(\ref{eq:asc_conv}) is further defined as:
\begin{equation}
\textbf{y}(\textbf{p}_0)=\sum_{\textbf{p}_n\in \mathcal{R}}\textbf{w}(\textbf{p}_n)\cdot \textbf{x}(\textbf{p}_0+{r(\textbf{x},\theta)}\cdot \textbf{p}_n)=\sum_{\textbf{p}_n\in \mathcal{R}}\textbf{w}(\textbf{p}_n)\cdot\sum_{\textbf{q}}f_{int}(\textbf{q},\textbf{p})\cdot\textbf{x}(\textbf{q}).
\label{eq:asc_conv_bilin}
\end{equation}
\subsection{Adaptive-scale convolutional neural network (ASCNet)}
\begin{figure}[!b]
	\centering
	\includegraphics[width=\textwidth]{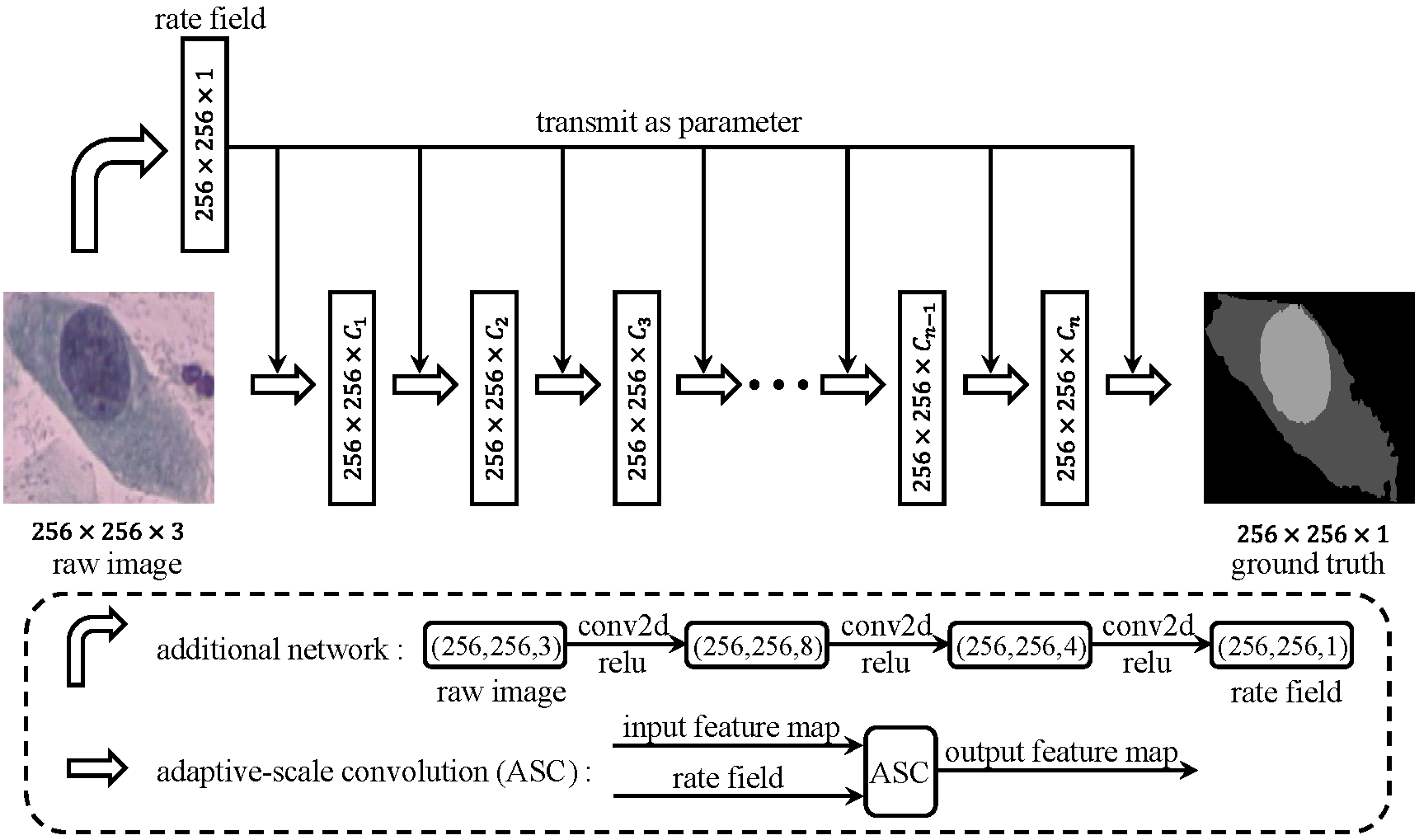}
	\caption{Architecture of the adaptive-scale convolutional neural network. Note that raw image, rate field and intermediate feature maps have the same spatial dimensions.}
	\label{fig:ASCNet_structure}
\end{figure}
Based on the ASC module, we present adaptive-scale convolutional neural network (ASCNet) to deal with pixel-wise semantic segmentation. As shown in Fig.~\ref{fig:ASCNet_structure}, ASCNet contains an additional network and a series of ASC modules connected in a sequential order. First, the dilation rate field is learned from the input image by an additional network which is formed by 3 $3\times3$ standard convolutions with the activation of RELU. We set the channel numbers of the 3 convolutions to 8, 4, and 1, respectively. Second, the learned rate field is transmitted to each ASC module, providing pixel-level dilation rates. The ASCs thus sample the multi-scale information with the optimal receptive fields as mentioned in Section 2.2. The proposed ASCNet is trained in an end-to-end fashion, on which the rate field and segmentation output are optimized simultaneously.
\section{Experiments}
\textbf{Data Description.} We evaluate the proposed method on two public datasets (The Herlev dataset \cite{jantzen2005pap} and SCD RBC dataset \cite{zhang2018rbc}): 1) The Herlev dataset consists of 917 cervical cell images (562 for train, 171 for valid and 184 for test) from Pap smear tests. All images are normalized to have zero mean with unit variance intensity and are resized to $256\times256$; 2) The SCD RBC dataset consists of 314 microscopy images (250 for train and 64 for test) from 5 Sickle Cell Disease (SCD) patients. Binary annotations (red blood cells VS background) are provided. We remove the blank margins of the raw images and resize them into $256\times256$. More information about the two datasets can be found in \cite{jantzen2005pap,zhang2018rbc}. \\
\\
\textbf{Experiments.} To further evaluate the effectiveness of the proposed ASCNet, we compare five models: 1) the classic CNN, which consists of 7 standard convolutions, and the channel numbers of the first 6 convolutions are set to 8 and that of the last layer is consistent with the number of class to generate the segmentation result; 2) the dilated CNN, which contains 7 dilated convolutions with the dilation rates of 1,1,2,4,8,16, and 1, respectively. Such setting follows the ``exponential expansion'' scheme  reported in \cite{yu2015multi}; 3) U-Net, the previous state-of-the-arts on these two datasets; 4) ASCNet-7, the proposed ASCNet with 7 adaptive-scale convolutions, while the channel number is regulated in the same way as the classic CNN; 5) ASCNet-14, the proposed ASCNet with 14 adaptive-scale convolutions, where the channel numbers of the first 13 convolutions are set to 32 and that of the last layer is set to the number of class. The kernel sizes are 3 for all convolutions in the experiments.\\
\\
\textbf{Implementation Details.} We train the models on a single NVIDIA GTX 1080ti GPU. To minimize the softmax cross entropy loss, networks are trained with Adam optimizer for 50000 epochs. Limited to GPU memory, each batch only contains one sample. Furthermore, we employ RELU as the activation function in all the convolutional layers (including the classic convolutions, the dilated convolutions and the ASCs). The networks are implemented in TensorFlow 1.2.1.
\begin{figure}[!t]
	\centering
	\includegraphics[width =\textwidth]{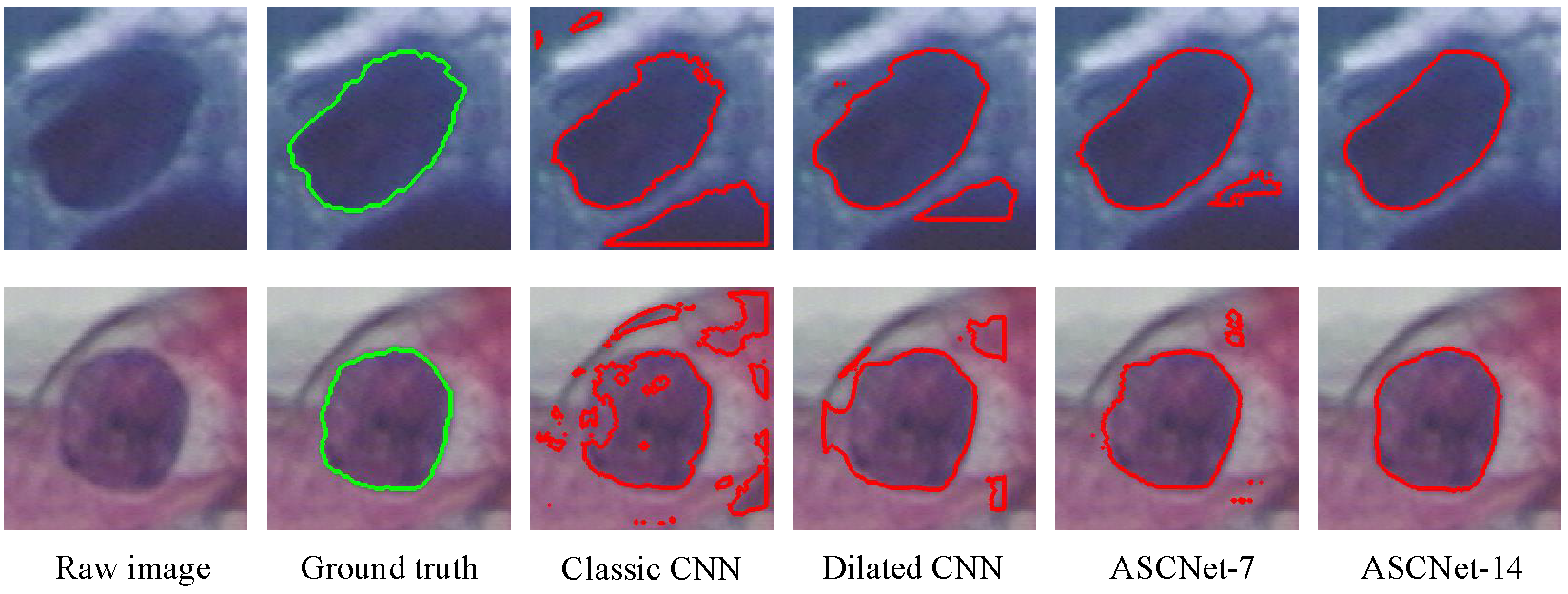}
	\caption{Two examples of the segmentation results of the different models on the Herlev dataset.}
	\label{fig:pap_segmentation}
\end{figure}
\section{Results}
\begin{table}[!b]
\begin{center}
\caption{Quantitative analysis of different methods in cell segmentation.}
\begin{tabular*}{0.72\linewidth}{c|ccc|ccc}
\hline
\multirow{2}{*}{}& \multicolumn{3}{c|}{The Herlev dataset} & \multicolumn{3}{c}{SCD RBC dataset} \\ 
& Dice & Precision & Recall & Dice & Precision & Recall\\ 
\hline
Classic CNN & 0.745   & 0.765 & 0.767 & 0.958   & 0.967 & 0.949 \\
Dilated CNN & 0.824   & 0.817 & 0.890 & 0.956  & 0.960 & 0.953 \\
ASCNet-7  & 0.857   & 0.863 & 0.891 & 0.959   & 0.960 & 0.958 \\
U-Net \cite{zhao2018automated,zhang2018rbc} & 0.869   & 0.897 & 0.879 & 0.957   & 0.955 & $-$ \\
\textbf{ASCNet-14} & \textbf{0.906}   & \textbf{0.909} & \textbf{0.925} & \textbf{0.967}   & \textbf{0.973} & \textbf{0.961} \\
\hline
	
\end{tabular*}
\label{tab:table1}
\end{center}
\end{table}
\textbf{Evaluation of the performance of ASCNet.} 
Table~\ref{tab:table1} reports the results of cell segmentation on the two datasets, where we use Dice, Precision, and Recall to measure the performance of five different approaches. 
The first three approaches, the classic CNN, the dilated CNN, and the proposed ASCNet-7, share the same network backbone of 7 convolutions with a kernel size of 3. As shown in Table \ref{tab:table1}, ASCNet-7 outperforms the other two, suggesting that the proposed ASC module is effective. Moreover, between the last two approaches, ASCNet-14 outperforms U-Net on both datasets, indicating that the proposed ASCNet can achieve accuracy comparable to that of the state-of-the-art methods. Lastly, compared to the ASCNet-7, ASCNet-14 benefits from the deeper (more convolutions) and wider (more channels) network architecture which is important to extract image information. 

Fig.~\ref{fig:pap_segmentation} illustrates two examples of the segmentation of different models on the Herlev dataset. It shows clearly that segmentation errors reduce significantly after we introduce ASC modules.\\
\\
\begin{figure}[!t]
	\centering
	\includegraphics[width =\textwidth]{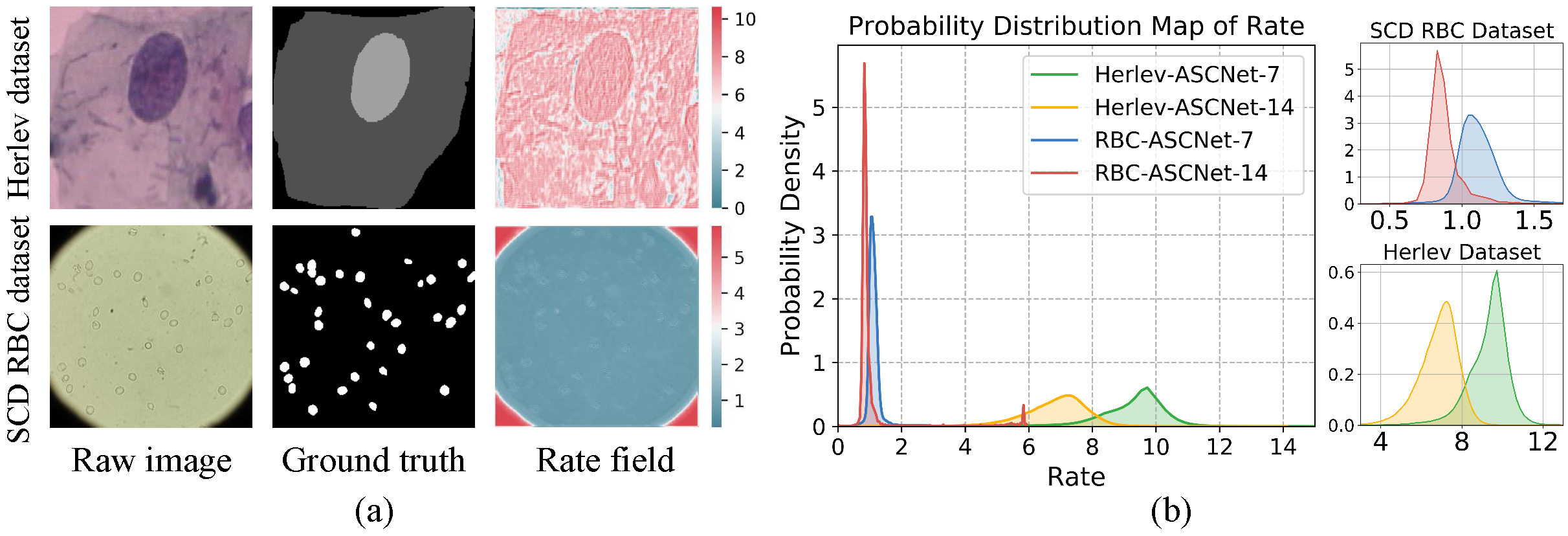}
	\caption{Visualization of the learned dilation rate. (a) Examples of the segmentation labels and learned rate fields. Top row: Herlev dataset; Bottom row: SCD RBC dataset. (b) An illustration of the distributions of the learned dilation rates. The red spike at approximately 6 is caused by the learned rates of image background in the four corners.}
	\label{fig:distribution_map}
\end{figure}
\textbf{Discussion about the learned dilation rate.} 
To better understand the mechanism of the ASC module, we display an example of rate fields generated by ASCNet-14 on both datasets in Fig.~\ref{fig:distribution_map} (a). Additionally, in Fig.~\ref{fig:distribution_map} (b), we display the corresponding distributions of learned dilation rates. For the Herlev dataset (top row in Fig.~\ref{fig:distribution_map} (a)), the cervical cells are imaged at high magnification. The ASCNet thus produces larger dilation rates accordingly, which are more capable of aggregating global information of the large objects. For the SCD RBC dataset (bottom row in Fig.~\ref{fig:distribution_map} (a)), the red blood cells are imaged at low magnification, so the ASCNet chooses to capture more local and fine details by producing smaller dilation rates. Another interesting finding is that the ASCNet can produce dilation rates less than 1, while the dilated convolution is incapable of doing so due to the integer dilation rate. In summary, the proposed ASCNet can adaptively learn the appropriate dilation rates for targets with different sizes, allowing the convolutions to better collect multi-scale features systematically. 

The dilation rates produced by ASCNet-14 are smaller than that produced by ASCNet-7 (see Fig.~\ref{fig:distribution_map} (b)). One possible explanation is that the final size of receptive field (accumulated by successive convolutional layers) required to extract information in a deep model is roughly constant for the same object. As all convolutions in ASCNet share the same dilation rate field, the receptive field in the deeper ASCNet is linearly magnified more times and each convolutional layer then requires smaller dilation rates. Such behaviors are worth further investigation with quantitative analysis in future work. 
\section{Conclusion}
In this work, we have proposed an adaptive-scale convolutional neural network (ASCNet) for image segmentation, which adaptively learns pixel-level dilation rates to form a dilation rate field, and ultimately, produces appropriate receptive fields for objects with different sizes. Experimental results demonstrate that the proposed ASCNet is effective in extracting multi-scale information and achieves the state-of-the-art accuracy on cell segmentation. Based on the visualization of the learned rate fields on the Herlev and the SCD RBC datasets, we discover a positive correlation between the optimal dilation rates and the sizes of the segmentation targets. This finding confirms that a key to improving multi-scale information extraction is to set the most appropriate receptive field for each object in the image. In addition, compared to its plain counterparts, ASCNet only introduces an additional 3-layer network with a slightly increased computational cost.  This allows us to easily replace the classic and dilated convolutions with ASCs in the existing deep models. In future work, we will test the proposed method on other computer vision tasks and further study the mechanism of multi-scale feature learning with ASC modules.

\bibliographystyle{splncs04}
\bibliography{reference}

\end{document}